\documentclass[11pt,parskip=half, a4paper]{article}

\usepackage{arxiv}

\usepackage[utf8]{inputenc} 
\usepackage{lmodern}
\usepackage[USenglish]{babel}
\usepackage[T1]{fontenc}    
\usepackage{hyperref}       
\usepackage{url}            
\usepackage{booktabs}       
\usepackage{amsfonts}       
\usepackage{nicefrac}       
\usepackage{microtype}      
\usepackage{lipsum}		
\usepackage{graphicx}
\usepackage{doi}
\usepackage{setspace}
\usepackage{subcaption}	
\usepackage{caption}
\usepackage{array,multirow,graphicx}
\usepackage{float}
\usepackage{color}
\usepackage[dvipsnames]{xcolor}

\usepackage[backend=biber,						
bibencoding=auto,
natbib=true,									
bibstyle=numeric,								
giveninits=true,									
citestyle=numeric,								
url=true,										
maxbibnames=50,								
minbibnames=3								
]{biblatex}
\bibliography{references}

\title{Utilizing the \textit{RRT*}-Algorithm for Collision Avoidance in UAV Photogrammetry Missions}

\author{Lars Killian  \\
	Technische Universität Braunschweig\\
	Institute of Flight Guidance\\
	Braunschweig, Germany \\
	\texttt{l.killian@tu-braunschweig.de} \\
	\And
	\href{https://orcid.org/0000-0001-8848-2144}{\includegraphics[scale=0.06]{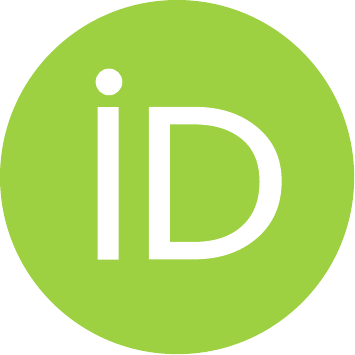}\hspace{1mm}Jan Backhaus} \\
	Technische Universität Braunschweig\\
	Institute of Flight Guidance\\
	Braunschweig, Germany \\
	\texttt{j.backhaus@tu-braunschweig.de} \\
}

\renewcommand{\undertitle}{}
\renewcommand{\shorttitle}{\textit{RRT*}-Algorithm for UAV Photogrammetry Missions}

\hypersetup{
pdftitle={Utilizing the \textit{RRT*}-Algorithm for Collision Avoidance in UAV Photogrammetry Missions},
pdfsubject={cs.RO, eess.SY},
pdfauthor={Lars Killian, Jan Backhaus},
pdfkeywords={To Do},
}

\begin{document}
\onehalfspacing
\maketitle

\begin{abstract}
This paper presents the application of the Rapidly-exploring Random Tree Star (RRT*) algorithm for multicopter collision avoidance in photogrammetry missions. For better applicability, the presented algorithm redirects the drone onto a predefined mission's path. The experiments are conducted in the simulation software gazebo utilizing a ROS interface to the widely known autopilot software PX4. For obstacle detection, a simulated Intel D435 stereo camera is used. The experiments include two different scenarios, each conducted with two different maximum velocities.  \newline
The results show that the probabilistic RRT*-algorithm can avoid obstacles successfully and intelligibly even at speeds up to 6 $\frac{m}{s}$. The main problems persist in the dynamic behavior, the inertia of the multicopter, and the limitations of the sensor technology. 
\end{abstract}

\keywords{UAV \and Multicopter \and Obstacle avoidance \and Rapidly-exploring-Random-Tree Star \and RRT* \and Photogrammetry }

\section{Introduction}
Unmanned aerial vehicles (UAVs) have seen an increase in popularity in recent years. Growing demand for autonomous task management requires solutions for numerous tasks such as obstacle avoidance. In addition to delivery and surveillance tasks, these vehicles are currently used for inspections. Camera-equipped UAVs are already used to determine damages, proportions and measurements of objects. One of these techniques is called photogrammetry. For this method, a UAV takes numerous uniformly spaced photographs, typically at a certain height. These missions often include several stripes and an inevitable overlap for each side of an image. \cite{linder2009} \newline
The advantages of a photogrammetry-mission are quick realization of a mission and  being able to utilize an afterwards developed computer model. When executed correctly, these models are highly accurate \cite{barry2013accuracy, aeroinspekt-mdpi}. With this method of inspection, construction progress, inaccessible or even dangerous areas can be mapped without endangering people in a safe and cost effective manner. Due to the previously described requirement of overlapping images, a redirecting of the UAV could benefit the resulting computer model when the UAV deviates from the predefined path to avoid obstacles. Although a well-predefined path would make this procedure obsolete, it could result in a higher time demand and, if done incorrectly, cause damages or endangerments. Additionally, moving objects would make this intention impossible. \newline
In this paper, we propose the RRT*-algorithm as a path manipulator. Thanks to the advantages like a fast calculation time, as demonstrated in \cite{peralta2020}, it can be used in real-world applications and complex environments. In addition, this algorithm evades the local minimum problem \cite{noreen2016optimal}. \newline
The algorithm is implemented into the PX4 framework and tested in the state-of-the-art simulation-software gazebo for evaluation. The obstacles are detected using a simulated intel D435 stereo camera. Subsequently, numerous tests are conducted in unknown space. When challenged with an obstacle, the RRT*-algorithm manipulates the path. After passing the obstacle, the UAV is redirected onto the predefined path to fulfill the survey-mission requirements. As most mapping cameras have no autofocus and adjusting the focal distance during a mission is generally impractical, our implementation of the algorithm is solely focused on generating avoidance trajectories in two dimensions.

\section{Rapidly-exploring Random Tree Star (RRT*) Algorithm}
\begin{figure}[h!]
 \centering
 \begin{subfigure}[b]{0.28\textwidth}
 	\centering
 	\includegraphics[width=\textwidth]{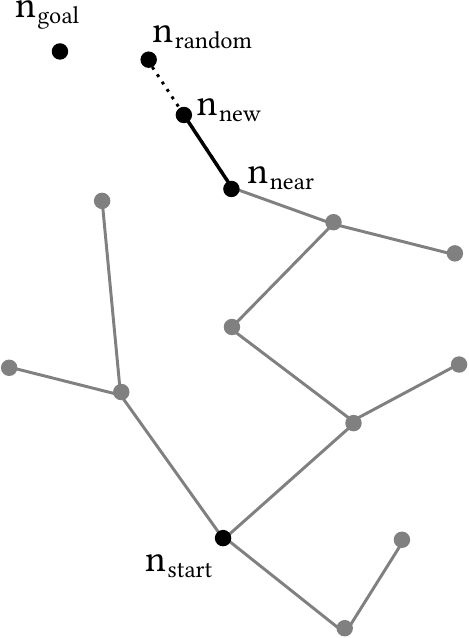}
 	\caption{RRT principle}
 	\label{RRT_principle}
 \end{subfigure}
 \hfill
 \begin{subfigure}[b]{0.28\textwidth}
	\centering 
	\includegraphics[width=\textwidth]{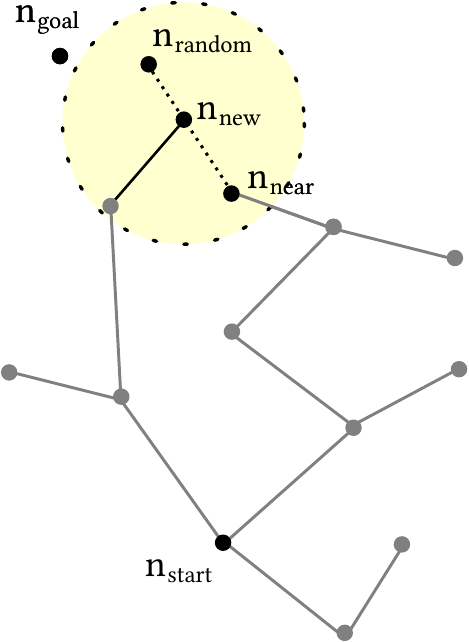}
	\caption{RRT*: \newline close proximity search}
 	\label{RRT-Star_principle}
  \end{subfigure}
 \hfill
 \begin{subfigure}[b]{0.28\textwidth}
	\centering 
	\includegraphics[width=\textwidth]{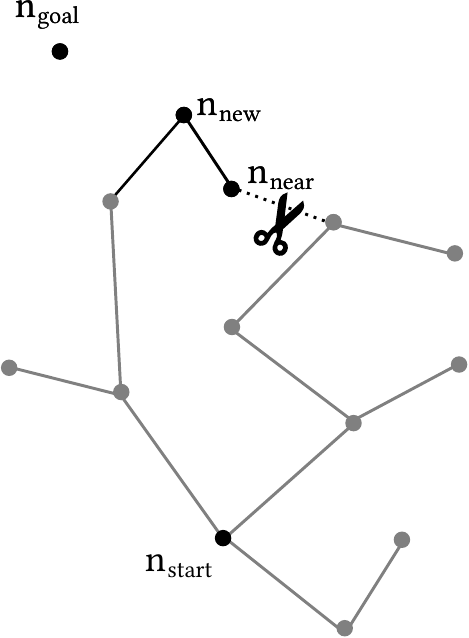}
	\caption{RRT*: \newline re-assignment of the parent node}
 	\label{RRT-Star_principle2}
  \end{subfigure}
  \caption{Rapidly-exploring Random Tree (RRT) and Rapidly-exploring Random Tree Star (RRT*) principle}
  \label{RRT_ges}
\end{figure}

The Rapidly-exploring Random Tree Star Algorithm, in short RRT*, is a modification of the Rapidly-exploring Random Tree. The probabilistic approach, first introduced by \cite{lavalle1998rapidly} in 1998, chooses nodes randomly and connects these with the closest node available. This simple procedure facilitates solving high-dimensional problems very quickly. \newline
As shown in Figure \ref{RRT_principle}, a starting node $n_{start}$ and a goal node $n_{node}$ are defined within a search space. A random node $n_{random}$ is generated from the free search space with each iteration, followed by a search for the nearest existing node $n_{near}$. Subsequently, on a virtual line connecting the newly created random node $n_{random}$ and the closest node $n_{near}$, a new node $n_{new}$ is created at a random distance from $n_{near}$. This distance is limited by the predefined maximum path resolution. In case the connecting line is collision-free, the node $n_{near}$ becomes $n_{new}$'s parent. The remaining node $n_{random}$ is deleted, and the next iteration step is performed. The algorithm ends as soon as a new node $n_{new}$ is within the goal node's path resolution radius and both nodes are connectable. \cite{tsardoulias2016} \newline
Due to its purely probabilistic behavior, the RRT algorithm often produces paths far from an optimal length. In contrast, the modified RRT* algorithm reduces the path length by considering several potential nodes before assigning the parent node. \newline
 \newline
As previously described, the traditional algorithm determines the new node $n_{new}$ on the connecting line between the randomly generated $n_{random}$ node and its closest node $n_{near}$. The modified RRT* algorithm now searches for other nodes within a search radius around $n_{new}$, as shown in figure \ref{RRT-Star_principle}. Next, the modified algorithm selects the parent node by its path length to reach $n_{new}$, not by the minimum distance. Ultimately, each path of the remaining nodes inside the search radius is compared to $n_{new}$'s path. When their length exceeds the sum of $n_{new}$'s length plus the distance between the considered node and $n_{new}$, $n_{new}$ takes its position as the parent node, thus disconnecting longer paths. This last step is visualized in figure \ref{RRT-Star_principle2}. \cite{tsardoulias2016}  \newline
Additionally to this procedure, the RRT* algorithm used for this paper randomly chooses the goal node and processes it as a randomly chosen node $n_{random}$ from time to time. Thus, generating branches striving towards the goal position. \cite{PythonRobotics}

\section{Implementation}
\begin{figure}[h!]
\centering
\includegraphics[scale=0.9]{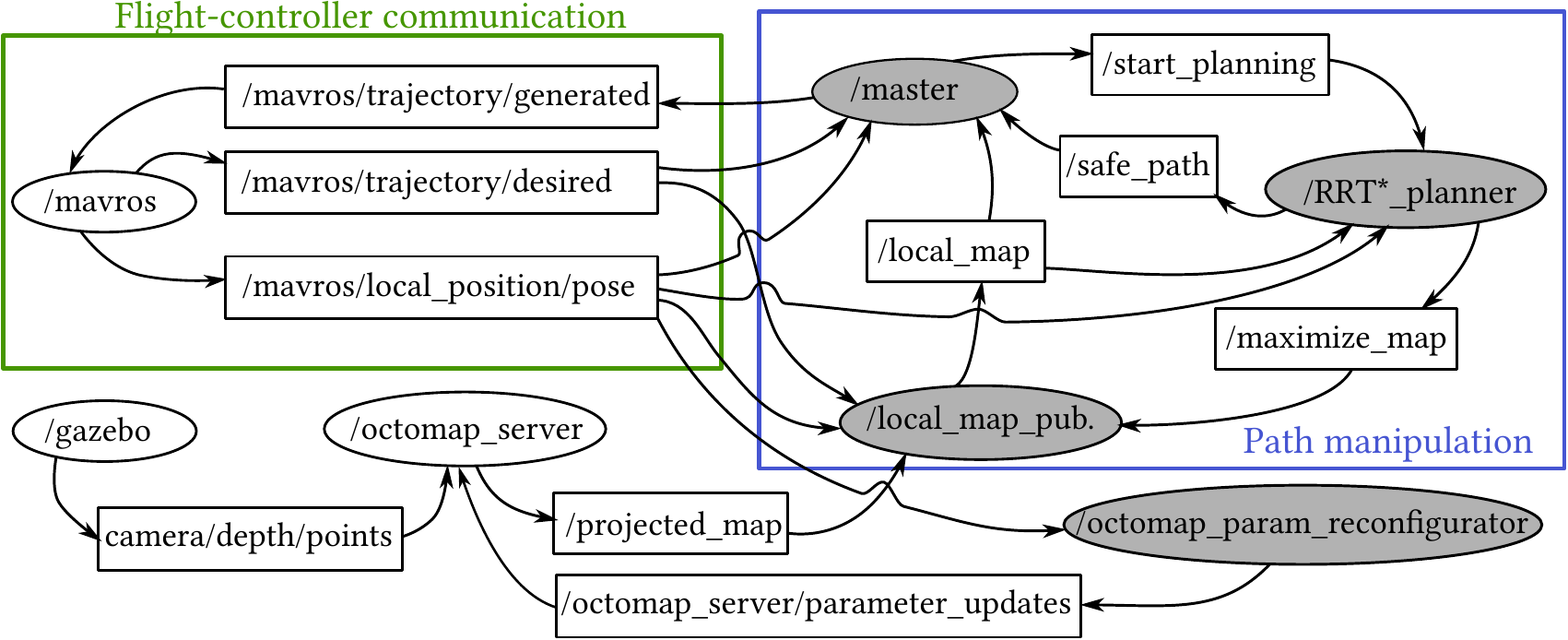}
\caption{Simplified node-topic relationship}
\label{RRT_implemented}
\end{figure}

The node-topic relationship that emerged during the implementation can be seen in Figure \ref{RRT_implemented}. Due to the general complexity of the whole graph, only the topics and nodes relevant for the procedure are visualized. The circled elements, also known as nodes, are responsible for the computation procedure and, if necessary, exchange messages using topics. A square box surrounds the latter. 
 \newline
To illustrate this research's important aspects, the graph is divided into sub-categories by using colored square boxes. Each of which will be explained in the following subsections.
\subsection{Flight-controller communication}
This subcategory is responsible for the communication between the flight controller and the companion computer. The flight-controller, using MAVLink messages, is not able to accept or even understand ROS-messages. This is where the mavros-node comes to hand. Mavros translates ROS-messages into MAVLink-messages and vice versa, serving as a communication driver for both parties. \cite{Mavros} \newline
The topics published by the mavros-node contain information regarding the drone's trajectory or position. To manipulate the drones desired trajectory, which is a point-by-point predefined path, the companion computer publishes points deviating from the original path.
\subsection{Simulation and Occupancy-mapping}
Outside the subcategories' definition, the so called \textit{gazebo}- and \textit{octomap\textunderscore server}-nodes are to be found. The point cloud generated by the virtual stereo camera in the simulation-software gazebo is transferred to the octomap\textunderscore server using the \textit{camera/depth/points}-topic. This topic holds information regarding occupied space in its close surroundings. With this data, a 3D occupancy grid map is constructed to represent the environment using the OctoMap framework \cite{Hornung2013}. It makes use of cubic volumes, called voxels, to represent occupied and free spaces. Due to its hierarchical data structure, the generated map keeps the required memory consumption to a minimum. In addition, the 3D environment is projected onto a two-dimensional grid map. This data is accessible via the \textit{projected\textunderscore map}-topic. \newline
During a phtogrammetry-mission, only the obstacles within the drone's altitude safety distance are considered dangerous to the drone. Therefore, only obstacles inside this range are taken into account for building the occupancy map. The minimum and maximum height range are determined by the client-node \textit{octomap\textunderscore pram\textunderscore reconfigurator}, which sends a request to the \textit{octomap\textunderscore server}-node to change its parameters accordingly.
\subsection{Path manipulation}
\begin{figure}[h!]
\centering
\includegraphics[scale=0.9]{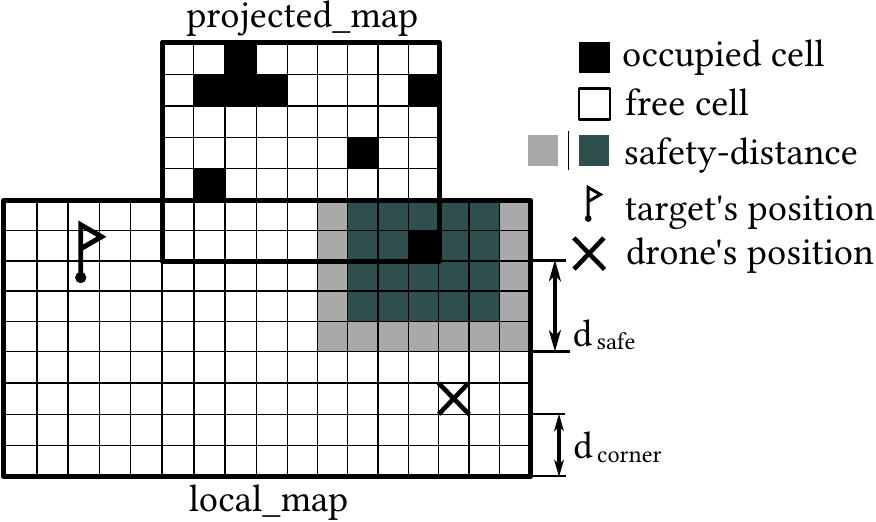}
\caption{Principle of the \textit{local\textunderscore map\textunderscore publisher}-node.}
\label{local_map_principle}
\end{figure}

\begin{figure}[h!]
\centering 
\begin{subfigure}{0.35\textwidth}
    \centering
    \includegraphics[width=\textwidth]{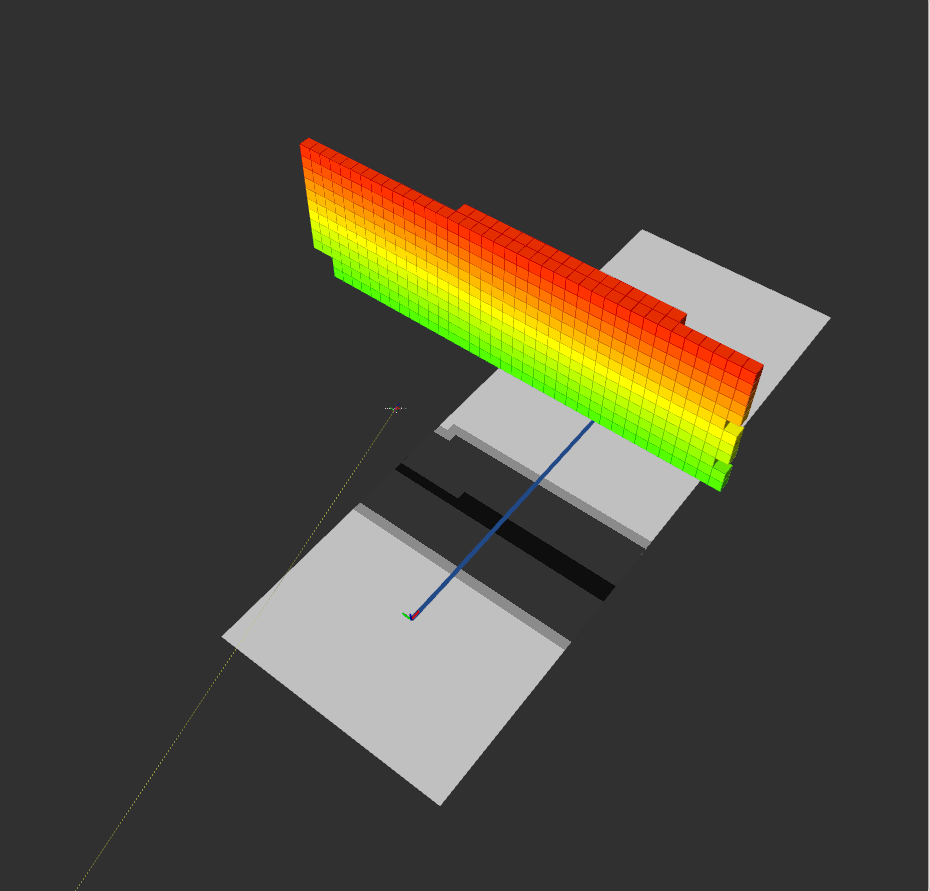}
    \caption{original $d_{corner}$} \label{small_map_rviz}
\end{subfigure}
\hspace{1em}
\begin{subfigure}{0.35\textwidth}
    \centering
    \includegraphics[width=\textwidth]{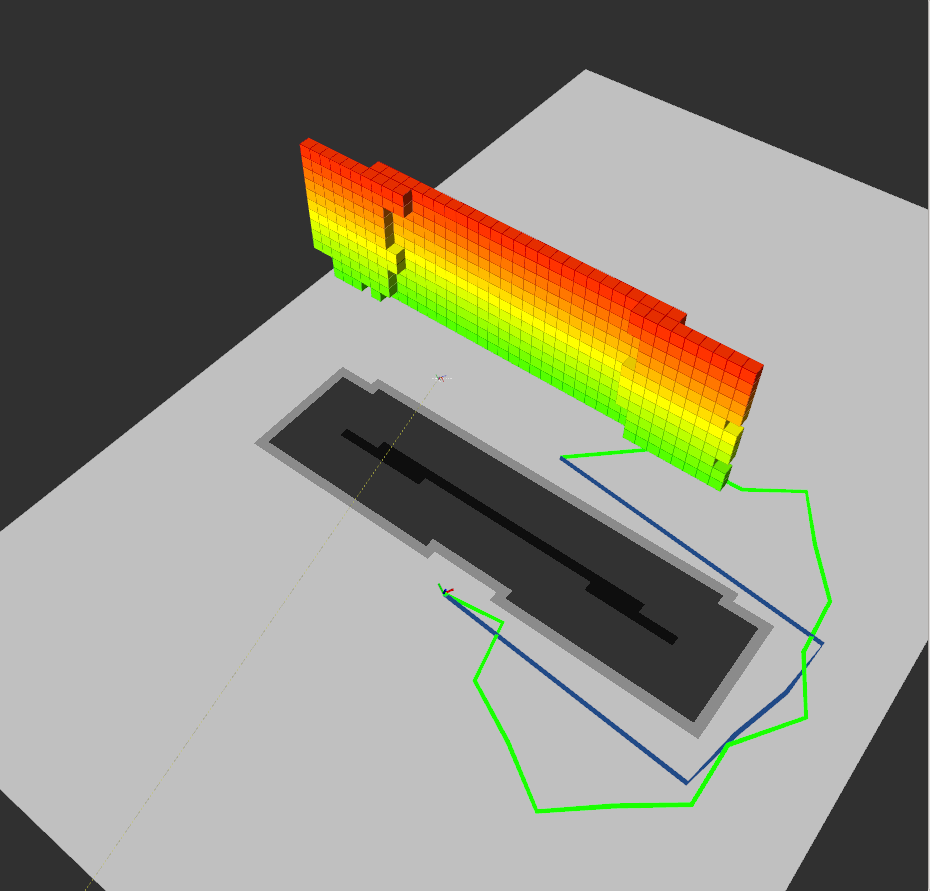}
    \caption{increased $d_{corner}$} \label{big_map_rviz}
\end{subfigure}
\caption{Effect of the \textit{maximize\textunderscore map}-request.}
\label{IncreaseMap_process}
\end{figure}

This subcategory illustrates the main implementation of the algorithm. The \textit{local\textunderscore map\textunderscore publisher} now uses the aforementioned \textit{projected\textunderscore map} in combination with the drone's position and the waypoint of the desired trajectory to create a grid map of its own. Figure \ref{local_map_principle} shows the created map. \newline
 \newline
In the first step, the node determines the area the drone will pass, therefore including the drone's and the target's position. Supplementary, a minimum distance $d_{corner}$ to the corners of the map is defined. The local map is thus reduced in size when closing in on the target position. Similarly, the map coverage is increased when the drone moves closer to the edge of the map than the the minimum corner distance $d_{corner}$. \newline
Next, the overlapping area of the \textit{projected\textunderscore map} is taken into account. When an occupied cell lays within this area, a safety distance $d_{safe}$ is created by occupying additional cells surrounding the obstacle. Due to the cells outside of the map, which are not taken into account, the corner distance $d_{corner}$ should be at least the size of the safety distance $d_{safe}$. Pictured as a slightly brighter outer rim of the safety distance contains a lower occupancy probability value, which will be explained later in this chapter. \newline
\begin{figure}[h!]
\centering
\includegraphics[scale=0.9]{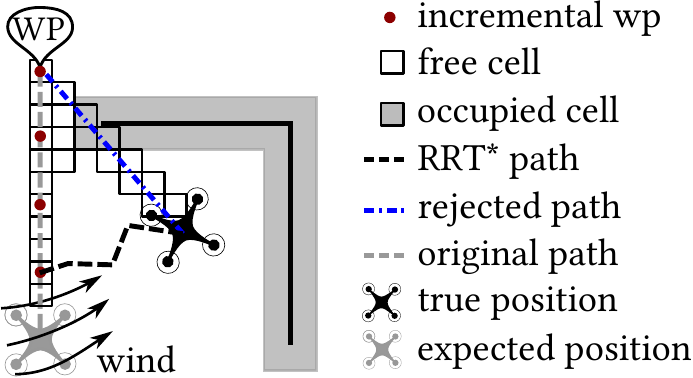}
\caption{Redirection of the drone towards the original path}
\label{master_path_checking}
\end{figure} \newline
After receiving the \textit{local\textunderscore map}, the \textit{master}-node undertakes multiple tasks beginning with subdividing the current mission path into incremental waypoints. If the drone later deviates from its path, it can return to the original predefined mission. Before publishing an incremental waypoint, the node verifies that the path is collision-free by assessing each passing cells' value. Should the drone be driven off course due to a wind gust, as shown in figure \ref{master_path_checking}, or the direct path is obstructed, the master-node sends a request to the \textit{RRT*\textunderscore planner}-node. Receiving the target position of an incremental waypoint that lays within a vacant cell, the RRT*-algorithm continuously starts planning possible paths, publishing these to the \textit{safe\textunderscore path}-topic until the drone reaches the desired incremental waypoint. \newline
Figure \ref{small_map_rviz} shows the potential risk of using a predefined map size. If the obstacle exceeds the map's borders, there is no possible solution for the RRT*-algorithm, resulting in an infinite calculation time. To reduce this risk, the RRT*-planner node can request an expansion of the map using the \textit{maximize\textunderscore map} command. Consequently, the \textit{local\textunderscore map}-publisher increases the minimum distance to the map's corners, as shown in figure \ref{big_map_rviz}. \newline
 \newline
The master-node now compares the incoming paths and decides which path to follow. One of the difficulties during trials to overcome was the repeatedly stopping of the drone in mid-air. The further exploration of the previously unknown obstacle caused the drone to reject its current trajectory and  wait for a new safe path. This usually resulted in an unnecessary wobbly behavior and sometimes a long waiting period. Instead of instantaneously rejecting the current path, the master node considers that the drone flies close to a potentially hazardous obstacle. The current path's cost:
\begin{equation}
C_{new} = C_{length} + n_{OR} \cdot cell_{size} \cdot 80 \%
\end{equation}
is calculated by the remaining path's length $C_{length}$, the number of cells passing the outer rim $n_{OR}$, the edge length of a cell $cell_{size}$, and a percentage value. The master-node compares this value with the incoming path lengths $C_{length}$ and then chooses which path to follow. \newline
An additional task of the master-node is the smoothing of the current RRT*'s-path. When the second node of the sequence is reachable by skipping the first node, the first node is deleted. This option is considered each time before the master-node publishes a new generated trajectory. While the \textit{local\textunderscore planner}-node cannot create paths inside the outer rim, the master-node can delete any node of the path, causing the drone to enter this area. Since the square-shaped safety distance causes the drone to choose longer paths around an obstacle, this behavior can be considered less dangerous than a continuous flight inside the outer rim.

\section{Experimental Setup}
Two different scenarios are considered for the experiments. In each scenario, an obstacle is placed on the mission's global path, blocking the direct path. The drone then starts from position (X=0, Y=0) and flies towards the goal (50,0). Along the way, a simulated Intel D435 stereo camera facing forwards is used to detect obstacles. According to it's specifications \cite{IntelD435}, a maximum sensor range of 10 meters is possible. The safety distance $d_{safe}$ is set to 3 meters.
\begin{figure}[h!]
\centering
\begin{subfigure}{0.325\textwidth}
    \centering
    \includegraphics[width=\textwidth]{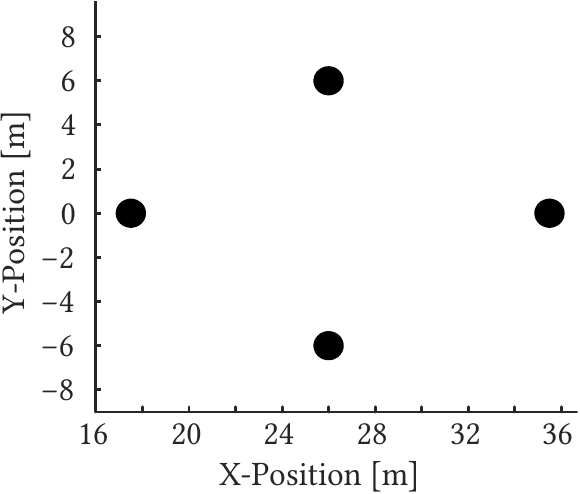}
    \caption{world 1 containing 4 pillars}\label{World_1_obs}
\end{subfigure}
    \hfill
    \begin{subfigure}{0.325\textwidth}
    \centering
    \includegraphics[width=\textwidth]{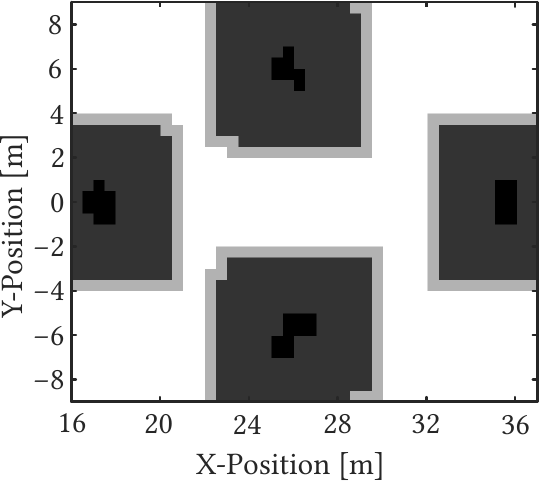}
    \caption{Resulting occupancy grid map for world 1}\label{World1_grid}
\end{subfigure}
\hfill
\begin{subfigure}{0.325\textwidth}
    \centering
    \includegraphics[width=\textwidth]{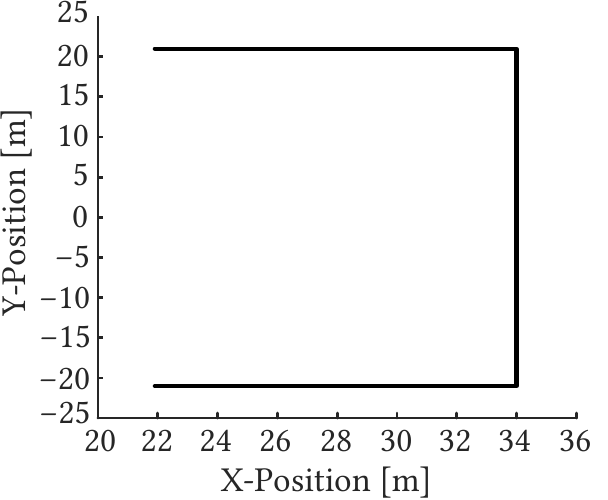}
    \caption{U-shaped obstacle facing towards the starting point}\label{World_2_obs}
\end{subfigure}
\caption{Obstacles of the simulated scenario}
\label{Preview_obstacles}
\end{figure} \newline
In the first scenario, two pillars are blocking the direct path, while an additional two are placed left and right, narrowing down the potential paths. This is shown in \ref{World_1_obs}. Due to the chosen safety distance, solely narrow passages are available to redirect the drone onto the predefined path. Figure \ref{World1_grid} illustrates a partial area of the generated local grid map.\newline
The second scenario intends to examine the drone's behavior in cases of significantly larger obstacles blocking its path. The continuous detection of the previously unknown area shall cause reoccurring rejections of previously chosen paths. This should provide information regarding the drone's dynamic behavior during this mission.  \newline
The simulation is carried out at two different velocities for each scenario. Furthermore, each velocity is tested ten times. The first velocity $V_{cruise}$ is set according to maximize the braking distance while at the same time the drone should still be able to avoid entering the safety distance. Using the values from table \ref{Table_PX4} and assuming that at the drone travels with the maximum velocity $V_{cruise}$ at the time of braking, this leads to an approximate velocity of 6 $\frac{m}{s}$. The second velocity is set to 4 $\frac{m}{s}$. 
\begin{table}[h!]
\centering
\def\arraystretch{1.5}
\begin{tabular}{|c|l|l|l|}
\hline
 Symbol& PX4 Parameter & Value & Description \\
 \hline
  $a'$ & MPC\textunderscore JERK\textunderscore AUTO & 4 $\frac{m}{s^{3}}$ & rate of change of acceleration \\
 $a_{max}$  & MPC\textunderscore ACC\textunderscore HOR& 5 $\frac{m}{s^{2}}$ &maximum horizontal acceleration \\
$V_{cruise}$ & MPC\textunderscore XY\textunderscore CRUISE & N/A & maximum horizontal velocity \\
\hline
\end{tabular}
\caption{PX4 parameters used for the simulation \cite{PX4Parameter}}
\label{Table_PX4}
\end{table} \newline
All simulations are conducted on a computer running the Ubuntu 18.04 operating system with the following specifications:
\begin{itemize}
\item CPU Intel Core i7-4790K @ 4.00GHz,
\item 24GB RAM memory.
\end{itemize}
The precise procedure to set up the simulation can be found in \cite{PX4Gazebo}. The scenarios and setup are available for download on GitHub at \cite{github-rrtstar}.

\section{Experimental Results}
\begin{figure}[h!]
\centering
\begin{subfigure}{0.45\textwidth}
    \centering
    \includegraphics[width=\textwidth]{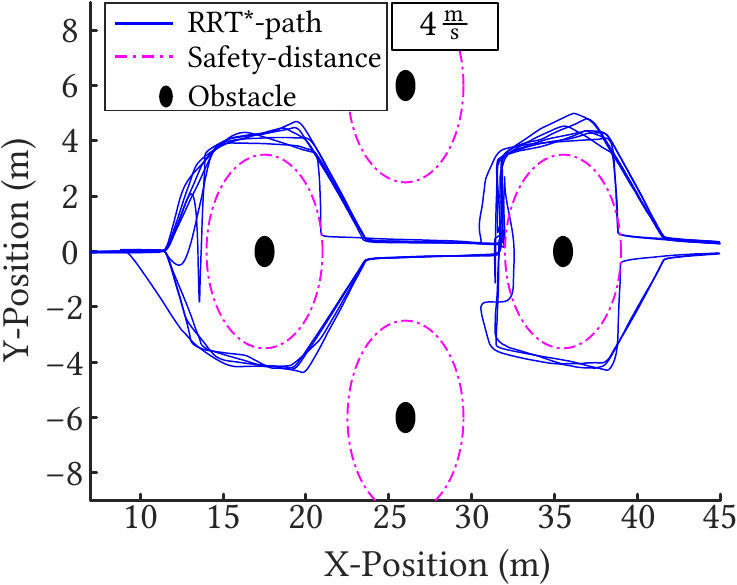}
    \caption{\large{$4 \frac{m}{s}$}}\label{World_1_4ms}
\end{subfigure}
    \hfill
\begin{subfigure}{0.45\textwidth}
    \centering
    \includegraphics[width=\textwidth]{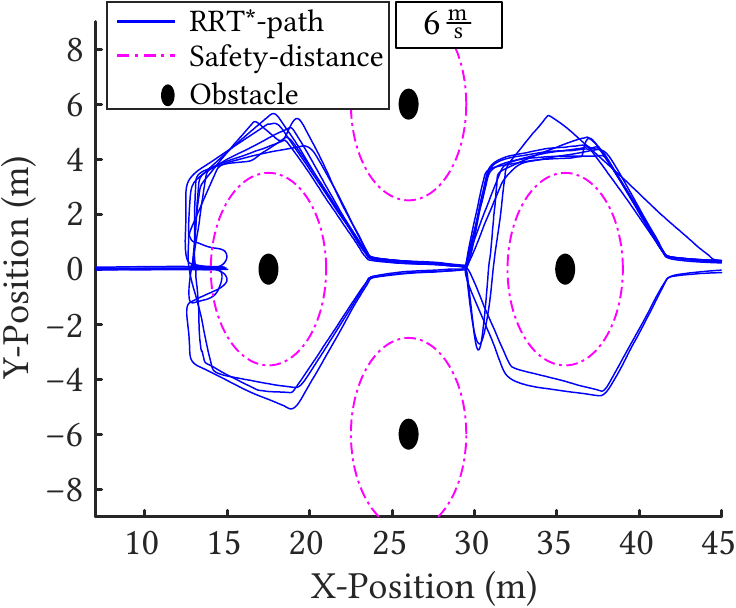}
    \caption{\large{$6 \frac{m}{s}$}}\label{world_1_6ms}
\end{subfigure}
	\bigskip
\begin{subfigure}{0.45\textwidth}
    \centering
    \includegraphics[width=\textwidth]{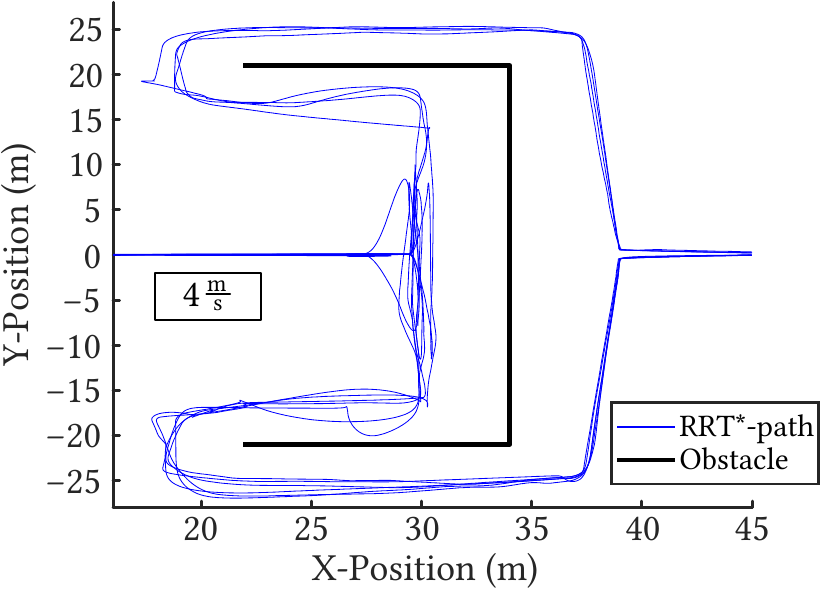}
    \caption{\large{$4 \frac{m}{s}$}}\label{world_2_4ms}
\end{subfigure}
\hfill
\begin{subfigure}{0.45\textwidth}
  \centering
  \includegraphics[width=\textwidth]{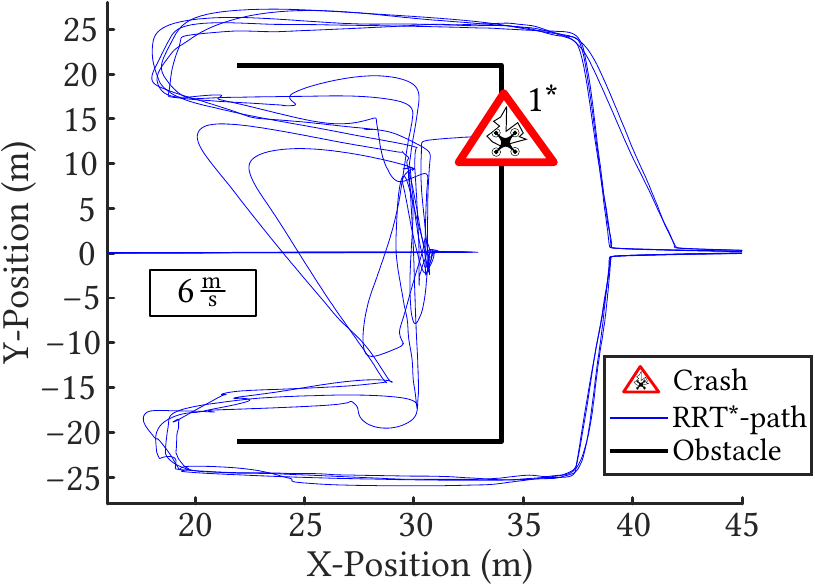}
  \caption{\large{$6 \frac{m}{s}$}}\label{world_2_6ms}
\end{subfigure} 
\caption{The course of all paths.}
\label{All_Path_figures}
\end{figure}

\begin{figure}[h!]
 	\centering
 	\includegraphics[scale=1]{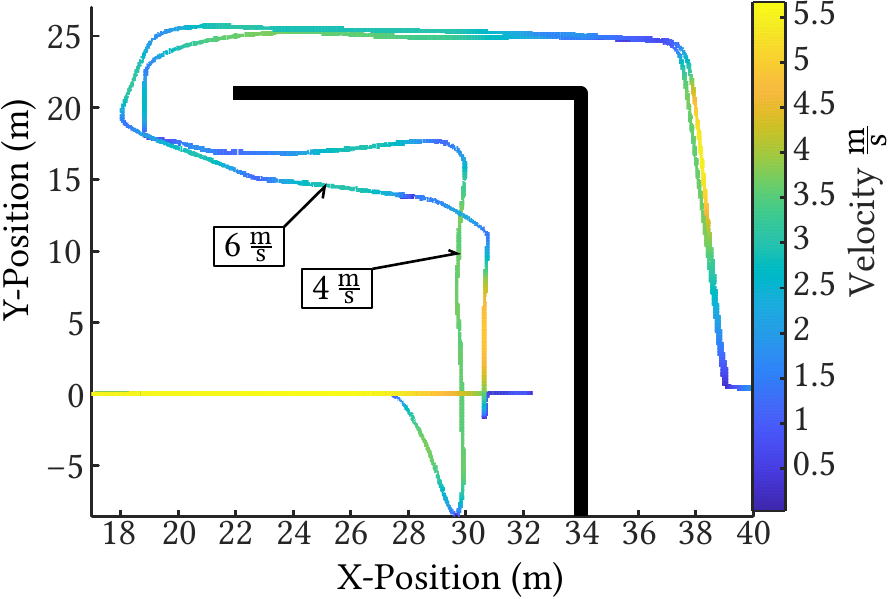}
 	\caption{Course of the velocity for $V_{cruise}$ = 4 $\frac{m}{s}$ and 6 $\frac{m}{s}$. Selected were the paths with the highest average velocity.}
 	\label{Velocity_Color_Map}
\end{figure}

The respective trajectories can be seen in figure \ref{All_Path_figures}. Initially, the drone avoids collisions successfully except for a single case. For seemingly obscure reasons, the drone crashed into the obstacle in figure \ref{world_2_6ms}. Due to the drone's unfortunate orientation, some parts of the obstacle were not fully detected. A resulting path directed the drone towards the assumed obstacle-free passage. The subsequent rotation did not occur fast enough to start the deceleration process, causing the unexpected interaction with the wall.
Tables \ref{Table_world_1} and \ref{Table_world_2} show the measured data regarding the maximum flown velocity $V_{MAX}$, the average velocity during the path planning $\bar{V}_{RRT^{*}}$, the traveled distance $D_{trv}$, and the average RRT*-rate $f_{RRT^{*}}$ in which a potential path was published. \newline
The average velocity $\bar{V}_{RRT^{*}}$ in scenario 1 hardly differs from the maximum cruise Velocity $V_{cruise}$. The reason for this is the waypoint. Paths within a higher occurrence of obstacles need an exact approach of the waypoint. Correspondingly, the flight controller is responsible to reduce the velocity to avoid an overshoot. All of the chosen paths follow the narrow passage between the obstacles. In scenario 2, the average velocity $\bar{V}_{RRT^{*}}$ does also not vary with the maximum cruise velocity $V_{cruise}$. Figure \ref{Velocity_Color_Map} shows an example of the course of the velocity during a single path for the respective maximum cruise velocity $V_{cruise}$. There are apparent differences in the peak velocity.
 \newline
In spite of the sudden appearance of an obstacle at the beginning of a mission, all paths avoid the collision successfully. The expected braking distance of 7m is exceeded by every single path for the higher cruise velocity. It proves to be the case that a crucial part of the obstacle avoidance system is the drone's dynamic attributes. Less agile and more sluggish drones would possibly collide with the obstacle. Furthermore, the RRT*-rate $f_{RRT^{*}}$ indicates that even higher velocities would be possible for the algorithm to operate successfully. Regardless of the drastic decrease in the rate for the second scenario, the published paths exceed 100 Hz by far. Numerous factors could cause this significant decrease: \newline
After the detection of the obstacle's appendage extending parallel to the x-axis (at Y = 21 or -21), the distance to the map's edge may again be too small to find possible solutions. This can be due to a complete blockage of the map, as at the beginning of the path's deviation, or simply because the algorithm does not find a possible solution within the maximum number of iterations—both resulting in an extended waiting period and a subsequent increase in the map size. In our measurement method, we can not distinguish these from one another and thus have to regard both as equivalent. Furthermore, it is unclear whether an increase in the maximum number of iteration would have led to a different outcome for the algorithm's possible tasks. The same applies to the number of iterations in which the goal node is selected as a random node.  Especially in the obstacle's inside, a larger number of paths result in a dead-end when a higher value is demanded. \newline
It should be noted however, that an increase in the map section size come simultaneously with an increase in computation time. Increasing the map's section size causes the algorithm to extend the following nearest node's search radius in addition to the number of cells to be checked for a collision-free path. \newline
\begin{table}[H]
\centering 
\bgroup
\def\arraystretch{2}
\begin{tabular}{|c|c|c|c|c|c|}
\hline
\multirow{2}{*}{ MPC\textunderscore XY\textunderscore CRUISE} &\multicolumn{2}{c|}{Veclocity in ($\frac{m}{s}$)} & RRT*-Rate in ($Hz$) & Path length in ($m$) & \multirow{2}{*}{ CMPL} \\
 & $\bar{V}_{MAX}$  $\pm \sigma$ & $\bar{V}_{RRT^{*}}$ $\pm \sigma$ & $f_{RRT^{*}}$ $\pm \sigma$  & $D_{trv}$ $\pm \sigma$  & \\
\hline
\large{4 $\frac{m}{s}$} & 3.81 $\pm$ 0.025 & 0.94 $\pm$ 0.53  & 459 $\pm$ 79.9 & 86.6 $\pm$ 5.4 & 10|10 \\
\hline
\large{6 $\frac{m}{s}$} &5.54 $\pm$ 0.04 & 0.97 $\pm$ 0.55 & 390 $\pm$ 110  & 84.7 $\pm$ 2.8  & 10|10 \\
\hline
\end{tabular}
\egroup
\caption{World 1: Expected values and standard deviations $\sigma$ }
\label{Table_world_1}
\end{table} 
\begin{table}[h!]
\centering 
\bgroup
\def\arraystretch{2}
\begin{tabular}{|c|c|c|c|c|c|}
\hline
\multirow{2}{*}{ MPC\textunderscore XY\textunderscore CRUISE} &\multicolumn{2}{c|}{Veclocity in ($\frac{m}{s}$)} & RRT*-Rate in ($Hz$) & Path length in ($m$) & \multirow{2}{*}{ CMPL} \\

 & $\bar{V}_{MAX}$  $\pm \sigma$ & $\bar{V}_{RRT^{*}}$ $\pm \sigma$ & $f_{RRT^{*}}$ $\pm \sigma$  & $D_{trv}$ $\pm \sigma$  & \\
\hline
\large{4 $\frac{m}{s}$} &  3.89 $\pm$ 0.01 & 1.98 $\pm$ 1.17 & 222 $\pm$ 132.7 & 117.4 $\pm$ 4.7 & 10|10 \\
\hline
\large{6 $\frac{m}{s}$} & 5.68 $\pm$ 0.004 & 1.93 $\pm$ 1.45 & 144 $\pm$ 61.4  & 122.4 $\pm$ 10.5 & 09|10 \\
\hline
\end{tabular}
\egroup
\caption{World 2: Expected values and standard deviations $\sigma$ }
\label{Table_world_2}
\end{table}

\section{Conclusion}
This paper shows that the implemented probabilistic RRT* path planning algorithm can be potentially used in UAV collision avoidance applications. The persistent difficulties lie in the limitations of the sensor technology and the dynamic behavior of the drone. Even though the drone crashed in the second scenario, we believe that a safe operation could have been guaranteed using additional stereo cameras on each side of the drone or by preventing the direction of flight from deviating from the direction of view. This would have prevented the formation of dead zones. Furthermore, the RRT*-algorithm showed that it has by far not reached its limitation. The limitations arose rather from the agility of the drone and sensor capabilities. \newline
The decision-making of the drones chosen path are occasionally surprising, especially the change of direction in scenario 1. An adjustment of the cost function, possibly by comparing the drones current orientation and velocity with the suggested path, could improve the current implementation.

\newpage


\printbibliography

\end{document}